%% file: main.tex
\documentclass[lettersize,journal]{IEEEtran}
\usepackage{amsmath,amsfonts,bm,mathtools}
\usepackage{algorithmic}
\usepackage{array}
\usepackage[caption=false,font=normalsize,labelfont=sf,textfont=sf]{subfig}
\usepackage{textcomp}
\usepackage{stfloats}
\usepackage{url}
\usepackage{verbatim}
\usepackage{graphicx}
\def\BibTeX{{\rm B\kern-.05em{\sc i\kern-.025em b}\kern-.08em
    T\kern-.1667em\lower.7ex\hbox{E}\kern-.125emX}}
\usepackage{balance}
\usepackage{xcolor}

\newcommand{\Var}{\mathrm{Var}}  

\usepackage{hyperref}

\usepackage{tabularx}
\usepackage{booktabs}

\usepackage{glossaries}

\loadglsentries[main]{glossary.tex}
\makeglossaries

\begin{document}

\title{Cryptocurrency Portfolio Optimization by Neural Networks}

\author{\IEEEauthorblockN{Quoc Minh Nguyen$^*$, Dat Thanh Tran$^*$, Juho Kanniainen$^*$, Alexandros Iosifidis$^\dag$, Moncef Gabbouj$^*$ \thanks{Q. M. Nguyen, D. T. Tran, J. Kanniainen and M. Gabbouj are with the Department of Computing Sciences, Tampere University, Tampere, Finland. (e-mail: \{quoc.nguyen,thanh.tran, juho.kanniainen, moncef.gabbouj\}@tuni.fi). A. Iosifidis is with the Department of Electrical and Computer Engineering, Aarhus University, Denmark (e-mail: ai@ece.au.dk).} }
\IEEEauthorblockA{\textit{$^*$Department of Computing Sciences, Tampere University, Finland}}\\
\IEEEauthorblockA{\textit{$^\dag$Department of Electrical and Computer Engineering, Aarhus University, Denmark}} \\
}

\maketitle


\begin{abstract}

Many cryptocurrency brokers nowadays offer a variety of derivative assets that allow traders to perform hedging or speculation. This paper proposes an effective algorithm based on neural networks to take advantage of these investment products. The proposed algorithm constructs a portfolio that contains a pair of negatively correlated assets. A deep neural network, which outputs the allocation weight of each asset at a time interval, is trained to maximize the Sharpe ratio. A novel loss term is proposed to regulate the network's bias towards a specific asset, thus enforcing the network to learn an allocation strategy that is close to a minimum variance strategy. Extensive experiments were conducted using data collected from Binance spanning 19 months to evaluate the effectiveness of our approach. The backtest results show that the proposed algorithm can produce neural networks that are able to make profits in different market situations.
\end{abstract}



\section{Introduction}

Portfolio optimization is the process of distributing wealth over a universe of assets to satisfy specific criteria (e.g., maximize accumulated return, Sharpe ratio, or minimize volatility). One of the pioneering works in this field is the modern portfolio theory \cite{markowitz_portfolio_1952}, which takes advantage of the diversification of the portfolio to reduce volatility, given the same level of portfolio returns. Recent research has incorporated machine learning into the model-based approach or designing a Reinforcement Learning (RL) framework in which an agent learns how to trade from market feedback, or even for training an end-to-end model to output the portfolio weights. Regarding different approaches, portfolio diversification still plays a vital role in the success of portfolio optimization algorithms. However, asset selection can be challenging since the universe of assets can be large and correlation properties between assets change over time.

In modern portfolio theory, the optimal weights are the solution to the optimization problem that uses the expected returns and covariance matrix as parameters. These two quantities are estimated from the historical returns, which have a low signal-to-noise ratio, and this is likely to affect the accuracy of the estimations. Many attempts have been made to bypass the need for the estimation step. One common approach is to design an RL framework \cite{filos_reinforcement_2019, jiang_deep_2017, wang_commission_2021} that uses an agent to take action based on market information (e.g., prices, returns, volumes) for maximizing the expected accumulated reward. For adapting the RL framework to the portfolio optimization problem, the action could be a vector of allocation weights the agent needs to allocate for constructing the portfolio. At the same time, the state can be created from the previous trading action, asset prices, and volumes. Another less-discussed but also promising approach is the end-to-end framework \cite{zhang_deep_2020, zhang_universal_2021, uysal_end--end_2021}. In this approach, the asset information forms an input to be introduced to a neural network which outputs the allocation weights. This paper will focus on the latter approach.

Choosing the market from which to construct the portfolio is also an important factor. Cryptocurrency is a type of virtual currency that uses cryptography technology so that it is not possible to make a counterfeit. The best-known example of cryptocurrencies is Bitcoin (BTC), which has a market capitalization of over 800 billion dollars\footnote[1]{Data reference from \url{https://coinmarketcap.com} on April 8, 2022} and is the highest market capitalization among other cryptocurrencies. In 2020, Binance, which is the largest cryptocurrency exchange in terms of the daily trading volume of cryptocurrencies, introduced Binance Leverage Tokens (BLVTs), with the first two of them called BTCUP and BTCDOWN. Those tokens are essentially tokenized versions of future positions of BTC allocated by Binance. This investment product lets traders perform hedging or speculate on BTC's future price movements (up or down) and achieve leveraged profits without liquidation risk.

This paper proposes to use an end-to-end machine learning framework to optimize the Sharpe ratio of a portfolio containing BTCUP and BTCDOWN. We estimate the price relation of the two assets and utilize this information to design a loss term that regulates the model’s bias towards a specific asset. When combined with the Sharpe ratio, this loss term can enforce the model to make allocations close to a neutral position and profit by taking arbitrage in these assets. In the ideal situation, the portfolio at the neutral position is invariant with respect to changes in BTCUP and BTCDOWN prices, and this position is constant over time. However, due to Binance’s proprietary allocation mechanism that affects the leverage effect and the trading activity of BLVTs on the market, this position changes all the time. Hence, our portfolio value can go up or down, even when we initially bought a portfolio of BTCUP and BTCDOWN at a neutral position.

This work makes several contributions. First, we study how to construct a portfolio from a pair of BLVTs. Second, we adopt an end-to-end machine learning framework to allocate a profitable BLVT portfolio in no-cost and cost-included settings. Finally, we make further improvements to the portfolio performance by exploiting the negative correlation characteristic of the BLVTs pair through a custom loss term. To the best of our knowledge, this is the first work that uses BLVTs for a portfolio optimization problem.



\section{Related Work}

Modern portfolio theory \cite{markowitz_portfolio_1952}, or the so-called mean-variance framework, is widely used to benchmark other portfolio optimization strategies. The goal is to maximize returns at a given risk level. Its variant, the Global Minimum Variance portfolio (GMVP) \cite{markowitz_portfolio_1952}, can be constructed by setting the objective function in the optimization step to the portfolio variance. This strategy is suitable for risk-averse investors whose focus is on reducing portfolio risk. Another common benchmark method is the capitalization-weighted portfolio \cite{hsu_cap-weighted_2004} which is used to build market indices such as S\&P500. In the capitalization-weighted approach, a large amount of money is distributed over a small set of high-capital assets, while the low-capital assets have little contribution to the portfolio performance. Hence, the diversification of the portfolio is decreased. This drawback is overcome by the Equal-weighted portfolio (EWP) \cite{demiguel_optimal_2009}, in which the total wealth is distributed equally to all assets. On the other hand, an equally-weighted risk contributions portfolio \cite{maillard_properties_2010} implements the idea of having the portfolio components contribute equal risk. Authors in \cite{maillard_properties_2010} state that this method offers a volatility level higher than the GMVP portfolio but lower than the EWP, allowing a trade-off between two approaches in terms of the absolute level of risk, risk budgeting, and diversification.

In the machine learning context, a natural approach for portfolio optimization is the RL framework since the problem involves interacting with a market via actions like asset allocation and received rewards such as portfolio returns. In the work of \cite{filos_reinforcement_2019}, the authors explored how the asset allocation problem can be addressed using the RL method. The experiments conducted with multiple trading agents showed superior results over some model-based portfolio optimization strategies. Deep Portfolio Management (DPM) \cite{jiang_deep_2017} introduces many improvements to the RL framework in the portfolio optimization problem. The most important proposal in this work is to use a neural network architecture that enables sharing of network weights between different assets. This approach allows the network to assess an asset based on patterns learned from other assets. Since the network weights are not specialized to any asset, this method can update the portfolio's constituents or even the size of the portfolio. Going further toward a real-world trading system, Hierarchical Reinforced trading system for Portfolio Management (HRPM) \cite{wang_commission_2021} is an RL framework that decomposes the trading process into hierarchical tasks of portfolio optimization and allocation or trading. In this framework, each task is associated with a separate policy. The high-level policy manages the portfolio optimization task that determines the portfolio weights at a lower frequency, such as days, while the low-level policy operates at a higher frequency and places limit orders to fulfill goals from the high-level policy.

In the RL frameworks, the objective function is an expected cumulative reward, whereas popular risk-adjusted metrics like the Sharpe ratio \cite{sharpe_sharpe_1994} cannot be decomposed into immediate rewards. To be able to use the Sharpe ratio as the optimization objective, the authors in \cite{zhang_deep_2020} proposed an end-to-end framework that directly outputs allocation weights from the asset prices and returns via a Long Short-Term Memory (LSTM). Risk-based allocation in \cite{uysal_end--end_2021} takes a similar approach. This method trained a fully connected neural network to output the allocation of risk contribution and solve the weight allocation from that risk contribution using an implicit optimization layer embedded in the neural network. The framework proposed in \cite{zhang_deep_2020} has the flexibility to be extended to incorporate different types of portfolio constraints such as cardinality, leverage, or maximum position for individual assets by utilizing several custom neural network layers. Overall, the end-to-end framework is simpler than the RL approach in terms of framework complexity, and it can be easier to design specialized components for a specific problem.

\begin{figure}[!t]
    \includegraphics[width=9cm]{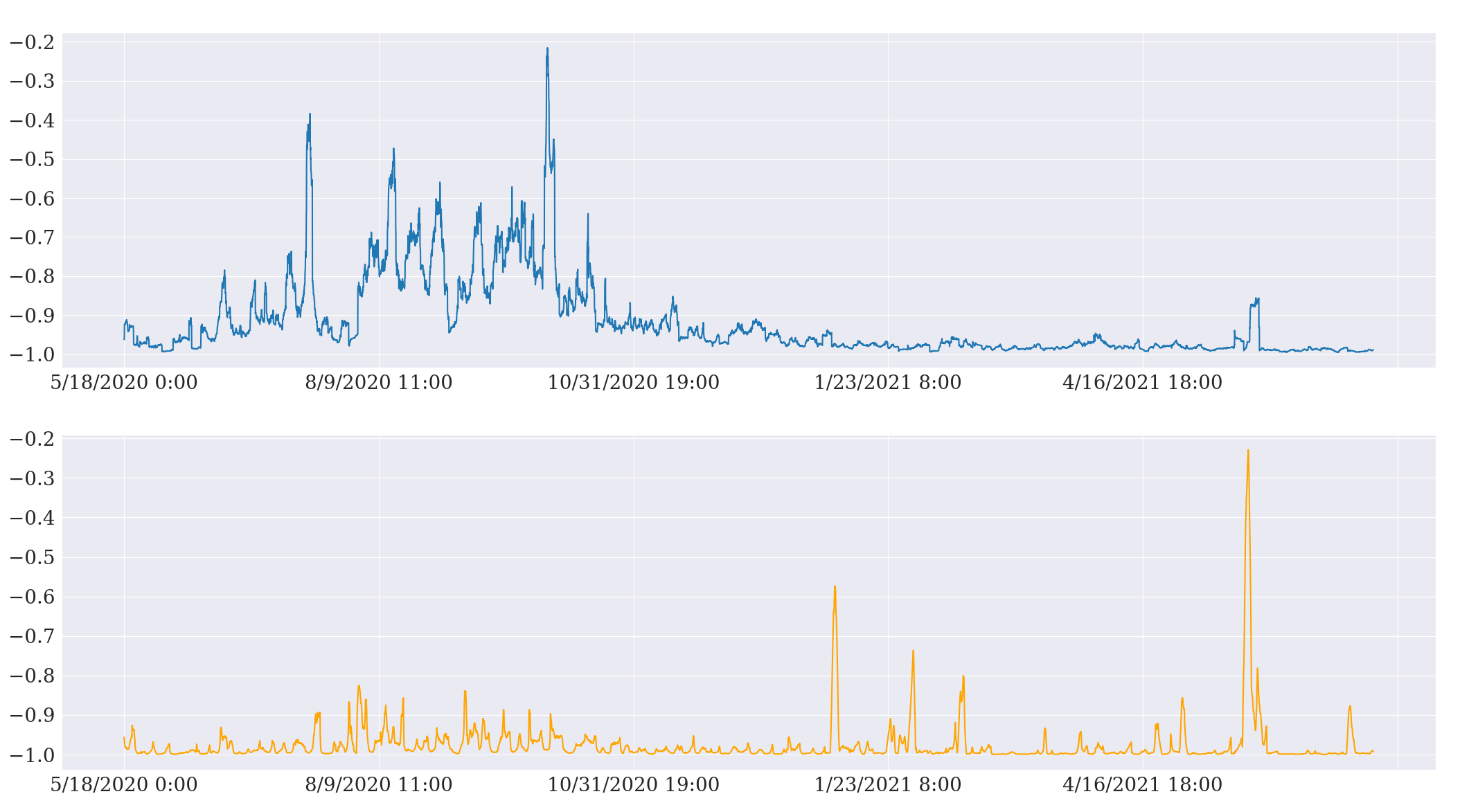}
    \caption{Rolling correlation of the returns (top) and the prices (bottom) of BTCUP and BTCDOWN from May 2020 to July 2021. The correlation window length is 72 hours, and the timeframe between price samples is 1 hour.}
    \label{fig:negative correlation}
\end{figure}

A special case of the portfolio optimization problem is the portfolio of Binance Leverage Tokens considered in this paper. Every BLVT is associated with a price called Net Asset Value (NAV). At this price, the owner can redeem their tokens to Binance and receive USDT\footnote[1]{USDT is a stablecoin. This type of cryptocurrency aims to keep cryptocurrency valuations stable. USDT pegs the US dollar, so the value of 1 USDT is very close to 1 US dollar.} back. The NAV is updated based on the fluctuation in the value of the future position basket that each BLVT represents. However, the BLVTs can also be traded in the spot market, where their prices follow around NAVs. These prices on the spot market also reflect speculation on the future price movements of the underlying assets in the futures market. The first available BLVTs are BTCUP and BTCDOWN. BTCUP aims to generate leveraged gains when the price of BTC goes up, while BTCDOWN aims to generate leveraged gains when the price of BTC goes down. From that property, every pair of UP and DOWN tokens in BLVTs have a consistent negative correlation over time. Figure \ref{fig:negative correlation} illustrates this correlation between BTCUP and BTCDOWN.



\section{Method}

This section describes the proposed strategy to construct a portfolio that consists of two leveraged tokens that are negatively correlated. In addition, this section presents the formulation of a novel loss function that is used to optimize a neural network generating the portfolio reallocation weights. The proposed loss function consists of two terms. The former maximizes the Sharpe ratio, while the latter is used to regularize the network's bias toward specific assets.


\subsection{Mathematical Formulation}
\label{mathematicalformalism}

Given an initial amount of capital, an agent allocates all capital into two assets, denoted as asset $A$ and asset $B$. At each time interval, the agent adjusts the amount of the two assets so that the portfolio contains only asset $A$ and asset $B$ (without any cash). The allocation of each asset in the portfolio is restricted to the long-only and budget constraints. The goal is to design a reallocation strategy that maximizes some portfolio metrics over time, for example, the Sharpe ratio. We start by constructing the mathematical formulation for a portfolio of $N$ assets as in \cite{jiang_deep_2017}. Later, we restrict $N=2$ for our specific problem. A portfolio of $N$ assets at time $t-1$ is associated with a weight vector
\begin{equation}
    \mathbf{w}_{t-1} = ( w_{t-1,0}, w_{t-1,1}, \dots, w_{t-1, N-1}  ),    
\label{eq:weight vector}
\end{equation}
which is the relative value of each asset compared to the portfolio value. At any time $t$, the long-only constraint is realized as
\begin{equation}
    w_{t,i} \geq 0 \quad \forall i \in \{0,1, \dots, N-1\},
\label{e:Long-only constraint}
\end{equation}
and the budget constraint can be written as
\begin{equation}
    \sum_{i=0}^{N-1} w_{t,i} = 1.
\label{eq:Budget constraint}
\end{equation}
Let denote period $[t-1,t]$ is the period between time $t-1$ and time $t$. At the beginning of period $[t-1,t]$, the asset price vector is denoted as
\begin{equation}
    \mathbf{y}_{t-1} = ( y_{t-1,0}\, , y_{t-1,1}\, , \dots, y_{t-1,N-1} ),
\label{eq:asset price vector}
\end{equation}
and the volume vector for the whole period is
\begin{equation}
    \mathbf{v}_{t-1} = ( v_{t-1,0}\, , v_{t-1,1}\, , \dots, v_{t-1,N-1} ).
    \label{eq:Voume vector}
\end{equation}
The asset value vector at time $t-1$, which denotes the absolute value (monetary value) of $N$ assets, is
\begin{equation}
    \mathbf{a}_{t-1} = \mathbf{y}_{t-1} \odot \mathbf{v}_{t-1}
    \label{eq:asset value vector at time t-1}
\end{equation}
where $\odot$ is the element-wise multiplication operator. The portfolio value is $p_{t-1}$, which is computed by summing over the absolute value of $N$ assets
\begin{equation}
    p_{t-1} = \mathbf{1}_N^T \, \mathbf{a}_{t-1} = \sum_{i=0}^{N-1} y_{t-1,i} \, v_{t-1,i},
    \label{eq:Portfolio value at time t-1}
\end{equation}
where $\mathbf{1}_N$ is the one-vector of size $N$.

At the end of period $[t-1,t]$, before we make any reallocation, the portfolio value is denoted as $p_{t}'$, and the asset price vector is $\mathbf{y}_{t}$. Here we assume that we only reallocate the portfolio at the end of each period, and asset prices at the beginning of the incoming period are equal to the prices at the end of the current period. The asset returns vector in period $[t-1,t]$ is
\begin{equation}
\begin{split}
    \mathbf{r}_t &= \left( ( \mathbf{y}_t - \mathbf{y}_{t-1} ) \oslash \mathbf{y}_{t-1} \right) \\
    &= \left( \frac{y_{t,0}}{y_{t-1,0}} - 1, \frac{y_{t,1}}{y_{t-1,1}} - 1, \dots, \frac{y_{t,N-1}}{y_{t-1,N-1}} - 1 \right),
\end{split}
\label{eq:asset_return}
\end{equation}
where $\oslash$ denotes the element-wise division operator. Right before making the reallocation in period $[t-1,t]$, the volume vector is determined by
\begin{equation}
    \mathbf{v}_{t-1} = p_{t-1} \mathbf{w}_{t-1} \oslash \mathbf{y}_{t-1},
\end{equation}
and the asset value vector is
\begin{equation}
   \mathbf{a}_t' = \mathbf{y}_t \odot \mathbf{v}_{t-1} = p_{t-1} \mathbf{w}_{t-1} \odot \left( \mathbf{y}_t \oslash \mathbf{y}_{t-1} \right).
\label{eq:asset vector value}
\end{equation}
Following Eq. (\ref{eq:asset_return}) and Eq. (\ref{eq:asset vector value}), the portfolio value at time $t$, before making reallocation, is
\begin{equation}
    p_t'= \mathbf{1}_N^T \, \mathbf{a}_t' = p_{t-1} \sum_{i=1}^{N-1} w_{t-1,i} (1 + r_{t,i}).
\label{portfolio_value_by_price_movement}
\end{equation}
The corresponding portfolio weight vector for this portfolio value is
\begin{equation}
    \mathbf{w}_t' = \frac{1}{p_t'} (\mathbf{y}_t \odot \mathbf{v}_{t-1}).
\end{equation}

\begin{figure}
    \centering
    \includegraphics[width=8cm]{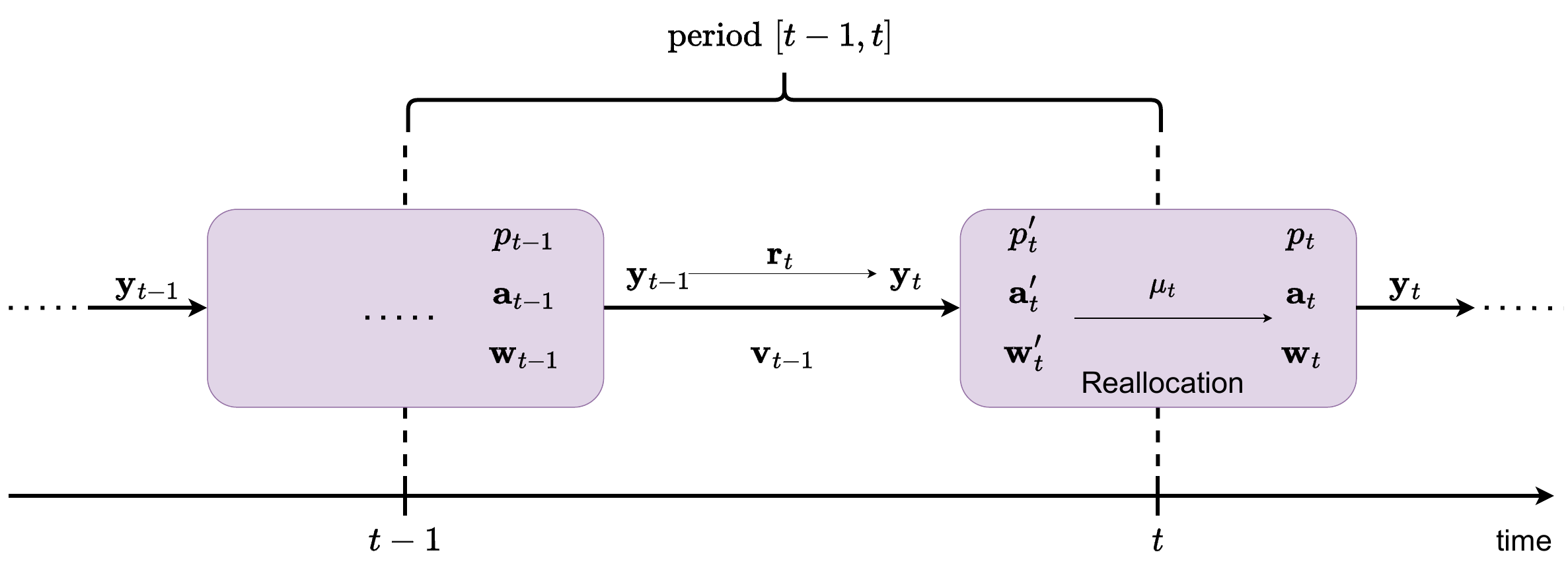}
    \caption{Illustration of the effect of price change and allocation to the portfolio quantities. In period $[t-1,t]$, the price of assets in the portfolio changes from $\mathbf{y}_{t-1}$ to $\mathbf{y}_t$, which results in the return vector $\mathbf{r}_t$. This price change adjusts $p_{t-1}$, $\mathbf{a}_{t-1}$, and $\mathbf{w}_{t-1}$ to $p_{t}'$, $\mathbf{a}_{t}'$, and $\mathbf{w}_{t}'$. The price change does not affect the volume vector $\mathbf{v}_{t-1}$. At the end of period $[t-1,t]$, the reallocation activity adjusts the previously mentioned quantities to $p_{t}$, $\mathbf{a}_{t}$, and $\mathbf{w}_{t}$. By the effect of transaction fee and other fees, the portfolio value $p_t'$ is shrunk to $p_t$ by a factor of $\mu_t$. By the reallocation, the volume vector has a new value.}
    \label{fig:mathematical formalism}
\end{figure}

At the end of period $[t-1,t]$, the agent needs to decide the weight vector $\mathbf{w}_t$ for the next period $[t,t+1]$ and makes the reallocation accordingly. This reallocation is executed via trading on the market, which requires trading fee. The cost reduces the portfolio value by a shrinkage parameter
\begin{equation}
    \mu_t = \frac{p_t}{p_t'},
\label{eq:mu_t_definition}
\end{equation}
where $0 < \mu_t \leq 1$. If there is no trading fee, the portfolio value will remain unchanged after the reallocation, which means $\mu_t$ is $1$. Figure \ref{fig:mathematical formalism} summaries the relationship between portfolio quantities in period $[t-1,t]$. 

The portfolio return in period $[t-1,t]$ is computed by comparing the portfolio value at the start of the current and incoming periods
\begin{equation}
    R_t = \frac{p_t}{p_{t-1}} - 1= \frac{\mu_t p_{t}'}{p_{t-1}} - 1= \mu_t \left( \sum_{i=0}^{N-1} w_{t-1,i} (1 + r_{t,i}) \right) - 1.
    \label{eq:portfolio return}
\end{equation}
The portfolio return of consecutive periods will be used to calculate the Sharpe ratio for training the neural networks model. This return is an essential input to compute other portfolio performance metrics.


\subsection{Transaction and Management Fee}
\label{transactioncost}

Transaction fee resulting from portfolio reallocation in the general case of $N$ assets has no analytical formula  \cite{ormos_performance_2013}, but it can be solved iteratively \cite{jiang_deep_2017}. In our portfolio optimization problem, there are only two assets, called $A$ and $B$, and cash is not allowed. When trading occurs, exactly one asset needs to be sold, and all the cash received afterward is used to buy the other asset. We can find the closed form of shrinkage parameter using this portfolio properties. 

At the end of period $[t-1,t]$, we need to adjust the weight vector from $\mathbf{w}_t'$ to $\mathbf{w}_t$ by reallocation. If the new weight for asset $A$ satisfied 
\begin{equation}
    w_{t,A}' \, p_t' > w_{t,A} \, p_t
    \Longleftrightarrow w_{t,A}' > \mu_t \, w_{t,A},
    \label{eq:Selling condition}
\end{equation}
then the allocation for this asset is decreased. In this case, we need to sell asset $A$, and the money obtained is used to buy more asset $B$. It can be shown that when the portfolio contains only two assets, the selling condition above is equivalent to $w_{t,A}' > w_{t,A}$. Similarly, asset $B$ is needed to buy more if and only if $w_{t,B}' < w_{t,B}$. 

We suppose that at the end of period $[t-1,t]$ the sold asset is asset $A$ and the bought asset is asset $B$.  When we perform buying and selling, these actions induce transaction fee. The change in the portfolio value before and after reallocation is caused only by these transaction fee. Therefore, the total trading fee from reallocation is $p_t' - p_t$. This cost can be broken down into the cost associated with selling and buying. We set the cost rate for selling and buying is equal and is denoted as $0 \leq c < 1$. The cost generated from selling asset $A$ is
\begin{equation}
    c \, (w_{t,A}' \, p_t' - w_{t,A} \, p_t) = c \, p_t' \, (w_{t,A}' - w_{t,A} \, \mu_t).
\label{eq:cost_generating_from_selling}
\end{equation}
The change in absolute value after reallocation of the bought asset $B$ is
\begin{equation}
    w_{t,B} \, p_t - w_{t,B}' \, p_t' = p_t' \, (w_{t,B} \, \mu_t - w_{t,B}').
\label{eq:change value of buying asset}
\end{equation}
However, to account for the trading fee, the buying order is larger than this value change by a multiple of $1/(1-c)$. Then, the cost generated from buying asset $B$ is
\begin{equation}
    c \, \frac{p_t' (w_{t,B} \, \mu_t - w_{t,B}')}{1-c}.
\label{eq:cost_generating_from_buy}
\end{equation}
The sum of selling cost and buying cost equals the difference between the portfolio values after and before reallocation
\begin{equation}
    c \, p_t' (w_{t,A}' - w_{t,A} \, \mu_t) +
    c \, \frac{p_t' (w_{t,B}\, \mu_t - w_{t,B}')}{1-c} =
        p_t' - p_t.
\end{equation}
Solving $\mu_t$ from this equation, the shrinkage parameter of the portfolio for reallocation is

\begin{equation}
    \mu_t = \frac{(1 - c) + c \, (w_{t,B}' - w_{t,A}' (1 - c))}
    {(1 - c) + c \, (w_{t,B} - w_{t,A} (1 - c))}.
\label{eq:mu_t formula}
\end{equation}
In case the sold asset is $B$, and the bought asset is $A$, we just need to exchange the index $A$ and $B$ of the weights in Eq. (\ref{eq:mu_t formula}) to obtain the right formula. 

BLVTs are also associated with the management fee. This fee is charged at 00:00 UTC directly on the net portfolio value. To consider this type of fee, we multiply $\mu_t$ by a multiplier $(1 - m)$ every $24$ hours, where $0 \leq m < 1$ is the management fee rate.


\subsection{Baseline Loss Function}
\label{sharperatio}

The Sharpe ratio integrates two important aspects of portfolio performance, that is, profitability and risk, into one measure. More specifically, it is defined as the excess portfolio expected return over the portfolio volatility. Since the portfolio return distribution is unknown, we estimate the Sharpe ratio with the portfolio return samples. For $T$ trading periods $\{[0,1], [1,2], \ldots, [T-1,T]\}$, the sample mean of portfolio return is  
\begin{equation}
    E_R = \frac{1}{T} \sum_{t=1}^T R_t,
\label{Expected return}
\end{equation}
where $R_t$ is determined by Eq. (\ref{eq:portfolio return}) and the portfolio volatility is computed by the sample standard deviation of portfolio returns, which  is
\begin{equation}
    \sigma_R = \sqrt{\frac{1}{T-1} \sum_{t=1}^T (R_t - E_R)^2}.
\label{Standard deviation}
\end{equation}
The Sharpe ratio, which omits the risk-free rate for simplicity, is
\begin{equation}
    SR_T = \frac{E_R}{\sigma_R}.
\label{eq:Sharpe ratio}
\end{equation}

In this paper, we train a neural network model, represented by the function $\mathcal{F}_{\bm{\theta}}(\cdot)$, to output the portfolio weights in an end-to-end manner \cite{zhang_deep_2020}. 
\begin{equation}
    \mathbf{w}_t = \mathcal{F}_{\bm{\theta}} (\mathbf{x}_t),
\label{eq:weight as model output}
\end{equation}
where $\mathbf{x}_t$ is the market information available at time $t$. The Sharpe ratio in Eq. (\ref{eq:Sharpe ratio}) is a function of portfolio weights. Therefore, it can be used as the loss function for training. We refer to this approach as the baseline method, which involves training a neural network model using the negative Sharpe ratio as the loss function.

\begin{equation}
    L_{BL} = -SR_T.
\end{equation}
Minimizing this loss function is equivalent to maximizing the Sharpe ratio. The long-only and budget constraints are fulfilled by adding the Softmax activation layer as the final layer in the neural network model.


\subsection{Neutral Position Constrain}
\label{Delta neutral loss term}

\begin{figure}
    \captionsetup[subfigure]{labelformat=empty}
    \centering
    \subfloat[]{\includegraphics[width=4.7cm]{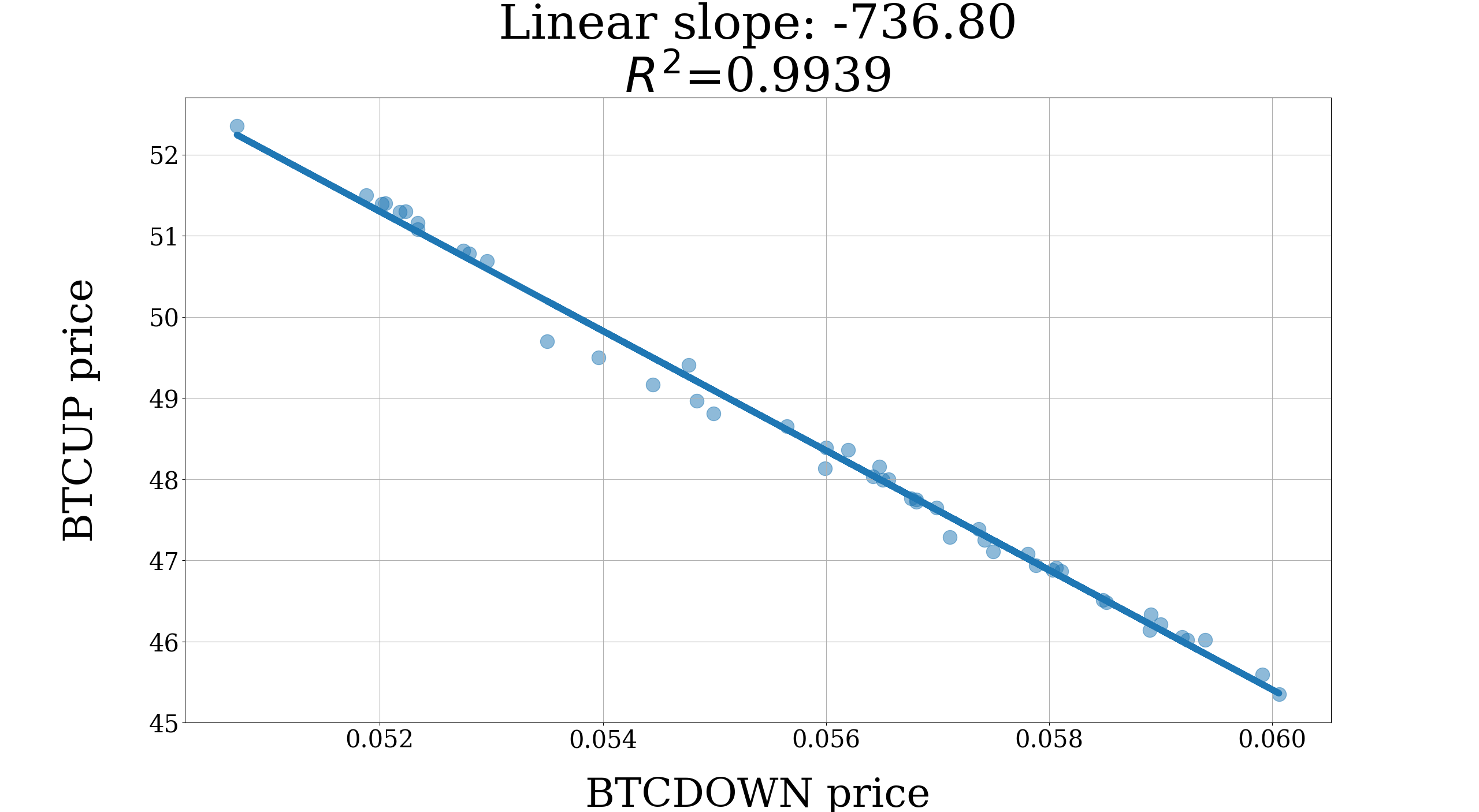}}\hspace*{-3mm}
    \subfloat[]{\includegraphics[width=4.7cm]{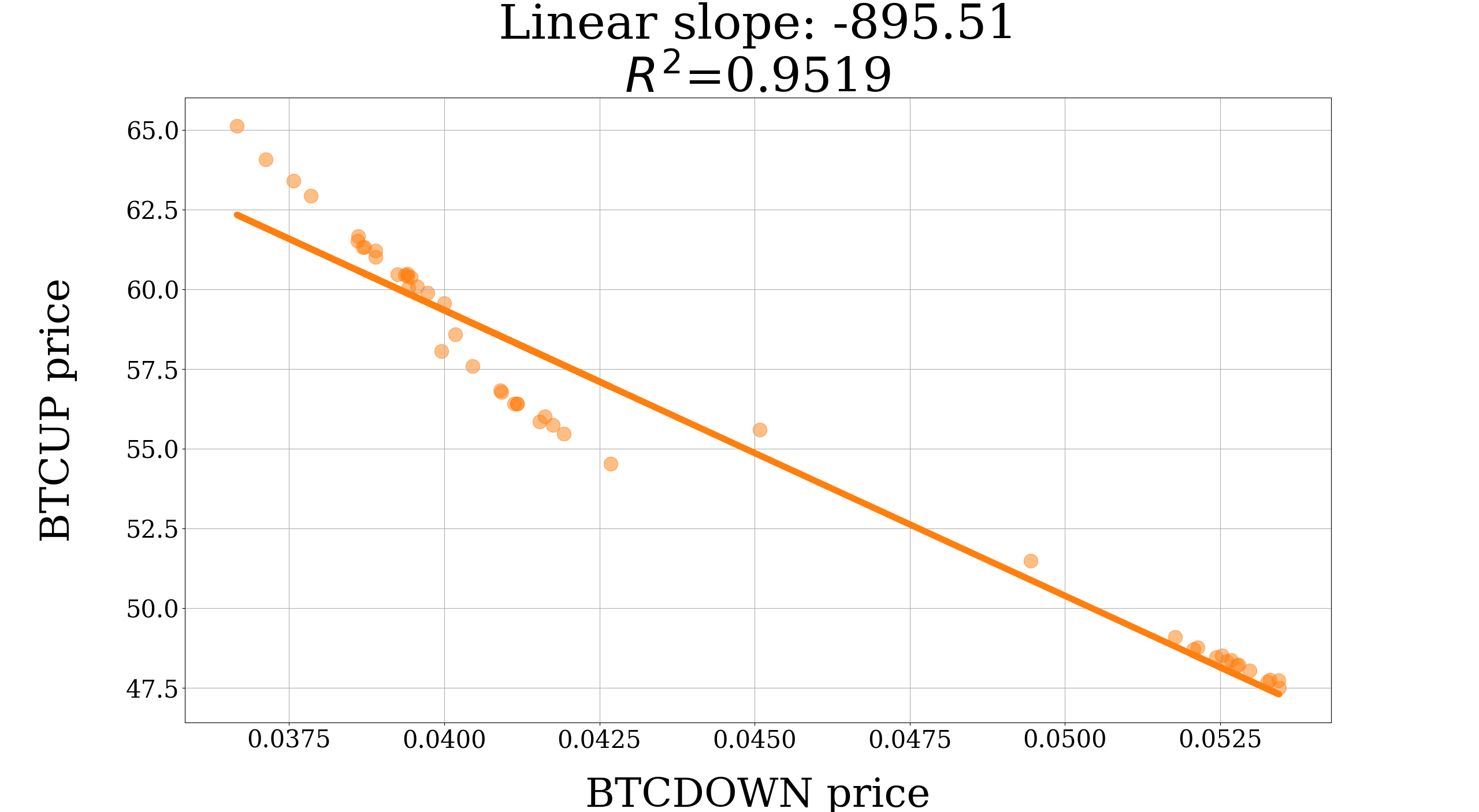}}\\[-3mm]
    \subfloat[]{\includegraphics[width=4.7cm]{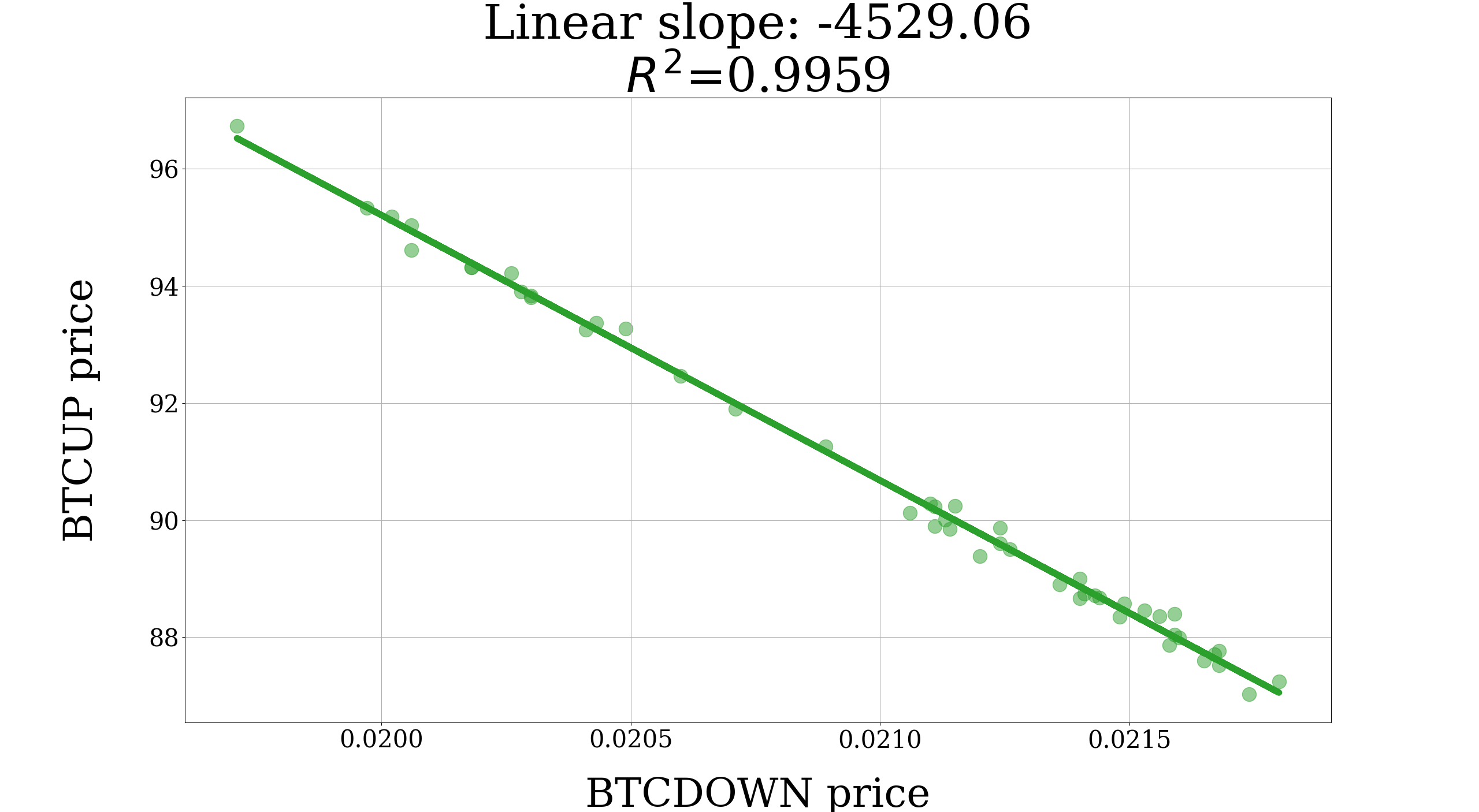}}\hspace*{-3mm}
    \subfloat[]{\includegraphics[width=4.7cm]{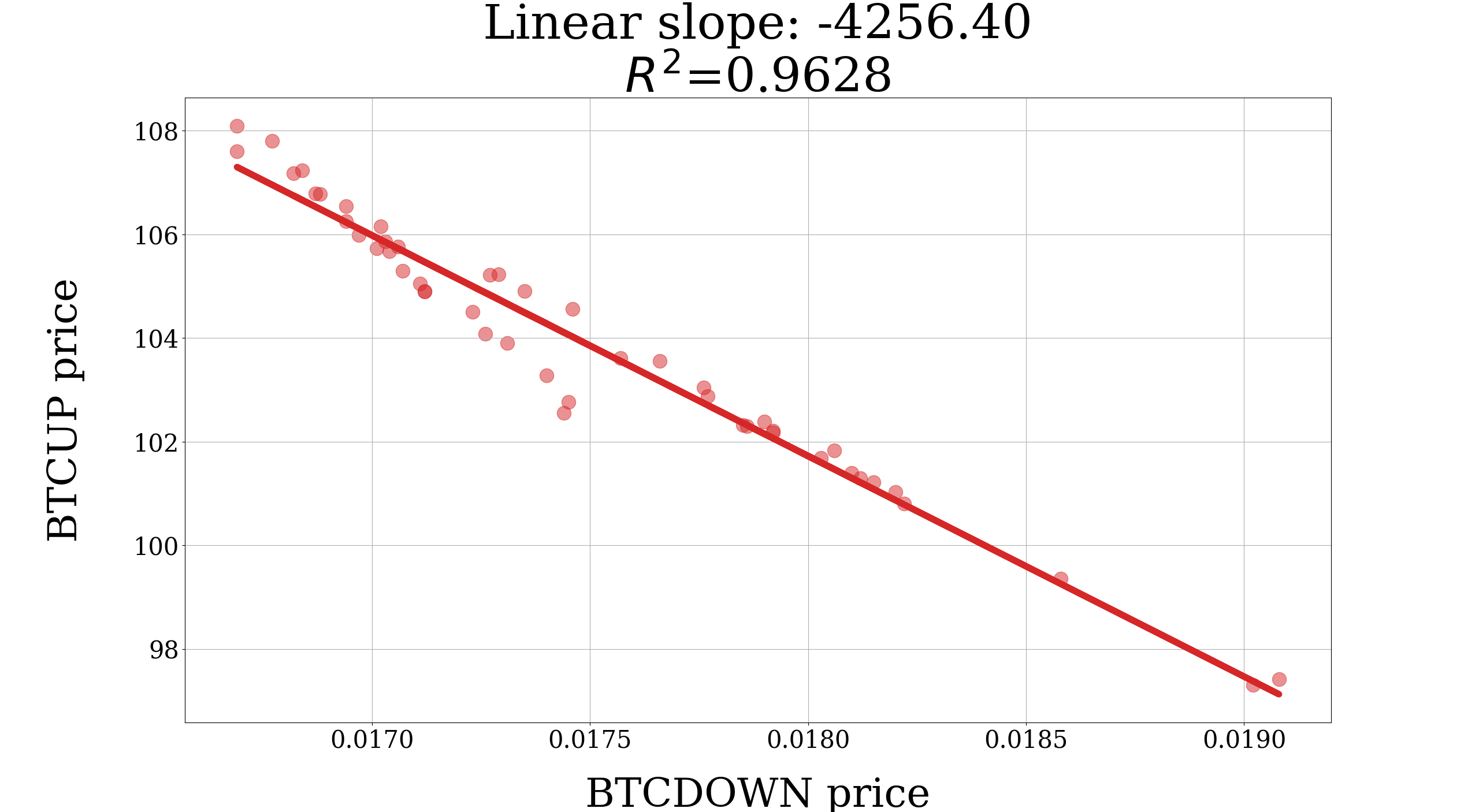}}%
    \caption{Scatter plot and ordinary least squares linear regression line of the prices of BTCUP and BTCDOWN. In each plot, the prices are sampled every $1$ hour and there are $48$ samples. Even though the price relation is a straight line, simple linear regression does not fit well in some cases.}
    \label{fig:Linear equations}
\end{figure}

In this section, we form a constrain for the training of neural networks for the optimal allocation of two tokens: BTCUP and BTCDOWN. This is based on linear regression to model the relationship between the price of the two tokens:
\begin{equation}
    y_{t,u} = \alpha + \beta^{market} \,y_{t,d} + \epsilon_t, 
    \label{eq:Line equation}
\end{equation}
The terms $y_{t,u}$ and $y_{t,d}$ denote the prices of BTCUP and BTCDOWN at time $t$, respectively. The error term $\epsilon_t$ is a zero-mean random variable with variance of $\sigma^2$. We assume that $\epsilon_t$ is independent with $y_{t,d}$. The coefficients $\beta^{market}$ and $\alpha$ are unknown parameters. Figure \ref{fig:Linear equations} illustrates the price relationship of two tokens and the corresponding estimated values of $\beta^{market}.$

Suppose that in period $[t,t+1]$, the price of BTCDOWN changes by $\Delta y_{t+1,d}$ to
\begin{equation}
    y_{t+1,d} = y_{t,d} + \Delta y_{t+1,d}.
    \label{eq:Down price in t+1}
\end{equation}
Then Eq. (\ref{eq:Line equation}) implies that 
\begin{equation}
    y_{t+1,u} = y_{t,u} + \Delta y_{t+1,d} \, \beta^{market} + ( \epsilon_{t+1} - \epsilon_t ).
\label{eq:Up price in t+1}
\end{equation}

At the beginning of period $[t,t+1]$, we hold a portfolio of BTCUP and BTCDOWN with volumes $v_{t,u}$ and $v_{t,d}\,$, respectively. The portfolio value at time $t$ is
\begin{equation}
    p_t = y_{t,u} \, v_{t,u} + y_{t,d} \, v_{t,d}.
    \label{eq:portfolio value of BTCUP and BTCDOWN at time t}
\end{equation}
The portfolio value at the end of the period $[t,t+1]$ just before the portfolio reallocation is
\begin{equation}
\begin{aligned}
    p_{t+1}' ={}& y_{t+1,u} \, v_{t,u} + y_{t+1,d} \, v_{t,d} \\
    ={}& p_t + (\beta^{market} \, v_{t,u} + v_{t,d}) \Delta y_{t+1,d} + v_{t,u} \, \Delta \epsilon_{t+1},
\end{aligned}
\label{eq:portfolio value in time t+1}
\end{equation}
where $ \Delta \epsilon_{t+1} = \epsilon_{t+1} - \epsilon_t$. 


Our regularization strategy is based on the idea that the value of the portfolio at $t+1$ just before the portfolio reallocation, $p_{t+1}'$, is {\em immune} for changes in $y_{t+1,d}$. In this way, our strategy is not dynamically dependent on instantaneous changes in the token prices, but is static. To exclude the second term in Eq. (\ref{eq:portfolio value in time t+1}), which captures the dependency on $y_{t+1,d}$, we set 
\begin{equation}
    \frac{v_{t,d}}{v_{t,u}} = -\beta^{market}.
\label{eq:Delta with mean}
\end{equation}
This does not guarantee that $p_{t+1}' = p_t$ as $\epsilon_{t+1} - \epsilon_t$ has certain variance, and for that reason, the portfolio value can change due to the residual component. However, assuming that $\epsilon_{t+1}$ does not correlate with $y_{t+1,d}$, this strategy minimizes the influence of $\Delta y_{t+1,d}$ on the $\Delta p_{t+1}' = p_{t+1}'- p_t $.

We call the weight values that satisfy Eq. (\ref{eq:Delta with mean}) the neutral weights and denote them as $w_{t,u}^*$, and $w_{t,d}^*$. Then by rewriting the volumes in terms of portfolio weights, we get
\begin{equation}
    \frac{w_{t,d}^*}{w_{t,u}^*} = -\beta^{market} \left( \frac{y_{t,d}}{y_{t,u}} \right).
\label{eq:Delta neutral weight}
\end{equation}
Suppose we explicitly set the portfolio weights to neutral weights in every period. In that case, the portfolio has no bias toward specific assets and is neutral regarding asset price change. One can expect the portfolio value to be stable in the successive trading period, and the maximum drawdown over the long term will be low. This strategy resembles the GMVP strategy because both aim to minimize portfolio volatility. However, when using the above analysis instead of GMVP, we directly exploit the linearity relation of a pair of BLVTs.

To compute the value of these neutral weights, we estimate the value of $\beta^{market}$ using the ordinary least squares method using data from the $K$ most recent period. Then
\begin{equation}
    \frac{\widehat{w}_{t,d}}{\widehat{w}_{t,u}} = -\widehat{\beta}_t^{market} \left( \frac{y_{t,d}}{y_{t,u}} \right),
    \label{eq:Delta-neutral weight estimation}
\end{equation}
where $\widehat{\beta}_t^{market}$ is the estimate at time $t$ of $\beta^{market}$, and $\widehat{w}_{t,d}$, and $\widehat{w}_{t,u}$ are the estimates of $w_{t,d}^*$, and $w_{t,u}^*$, respectively. Eq. (\ref{eq:Delta-neutral weight estimation}), together with the budget constraint, determines the value of $\widehat{w}_{t,u}$, and $\widehat{w}_{t,d}$.


\subsection{Variance-controlled Loss Terms}
\label{Delta controlled loss term}

Controlling the neutrality of a portfolio has some benefits. First, gearing the portfolio towards neutral position will lower directional risk, hence possibly improving the Sharpe ratio. Second, reducing the network's bias toward a specific assets can reduce portfolio lost in case the asset has worse performance compare to its past. Finally, the region around the neutral weights by a margin will always contain some weight values that the model can make profits or at least preserve the portfolio value. If we encourage model outputs to be in this region, it might be easier for the network to learn how to take profits rather than searching from random portfolio weight values.


To control the neutrality of the portfolio, we first rewrite the approximate portfolio variance
\begin{equation}
\begin{split}
    \Var[p_{t+1}'] &\approx ( \beta^{market} \, v_{t,u} + v_{t,d} )^2 \, \Var[\Delta y_{t+1,d}] \\
    &= v_{t,u}^2 ( \beta^{market} - \beta_t^{model} )^2 \, \Var[\Delta y_{t+1,d}],
\end{split}
\end{equation}
where
\begin{equation}
    \beta_t^{model} = -\frac{v_{t,d}}{v_{t,u}} = -\frac{y_{t,u} \, w_{t,d}}{y_{t,d} \, w_{t,u}}
\label{eq:Delta model definition}
\end{equation}
Here $\beta_t^{model}$ is a parameter computed from model allocation weights. At training time, we use the estimate $\widehat{\beta}_t^{market}$ for $\beta^{market}$. Our purpose is to constrain the variance via the term $(\beta_t^{market} - \beta_t^{model} )^2$ within a margin
\begin{equation}
\begin{split}
    (\beta_t^{market} - \beta_t^{model} )^2 &\leq ( \gamma \beta^{market} )^2 \\
    \Longleftrightarrow (1 + \gamma) \beta^{market} \leq \beta_t^{model} &\leq (1 - \gamma) \beta_t^{market}.
\label{eq:Delta bound conditions}
\end{split}
\end{equation}
This condition is equivalent to
\begin{equation}
\begin{cases}
    C_1 =\beta_t^{model} - (1 + \gamma) \beta_t^{market} \geq 0 \\
    C_2 = (1 - \gamma) \beta_t^{market} - \beta_t^{model} \geq 0.
\end{cases}
\end{equation}
where $ \gamma \geq 0 $ is the parameter that controls the degree of neutrality of the portfolio. 
We design the following loss terms that take the form of hinge loss
\begin{equation}
    A_1(\mathbf{w}_t;\gamma) = \max(0, -C_1 \, C_2),
    \label{eq:A_1}
\end{equation}
and
\begin{equation}
    A_2(\mathbf{w}_t;\gamma) = \max(0,-C_1)^2 + \max(0, -C_2)^2.
    \label{eq:A_2}
\end{equation}
These two loss terms will only penalize the model when the $\beta_t^{model}$ lies outside the constrained region. When $\beta_t^{model}$ lies in the range defined in Eq. (\ref{eq:Delta bound conditions}), the two loss terms will vanish. Otherwise, the penalized terms will be equal to the following positive terms
\begin{equation}
    A_1(\mathbf{w}_t; \gamma) = -C_1 C_2, \\
\end{equation}
and
\begin{equation}
    A_2(\mathbf{w}_t; \gamma) = 
    \begin{cases}
    C_1^2 & \text{if $C_1 < 0$}, \\
    C_2^2 & \text{if $C_2 < 0$}.
    \end{cases}
\end{equation}

We have conducted experiments where we train the neural network with loss functions that is the negative Sharpe ratio combined with $A_1(\mathbf{w}_t;\gamma)$ or $A_2(\mathbf{w}_t;\gamma)$. The obtained Sharpe ratios are significantly lower than the baseline. We suspect that the model output in a fraction form $w_{t,d}/w_{t,u}$ in the proposed loss terms may cause difficulty for the model to learn the optimal allocation. We solve this problem by multiplying the proposed loss terms with square BTCUP volume $v_{t,u}^2$ to eliminate the fraction form and observe better results. The final forms of the proposed loss terms are
\begin{equation}
\begin{split}
     HL_1(\mathbf{w}_t; \gamma) &= v_{t,u}^2 \, A_1(\mathbf{w}_t; \gamma) \\
            &= \max \left(0, - v_{t,u}^2 \, C_1 \, C_2 \right),
\end{split}
\label{eq:HL1}
\end{equation}
and
\begin{equation}
\begin{split}
     HL_2(\mathbf{w}_t; \gamma) &= v_{t,u}^2 \, A_2(\mathbf{w}_t; \gamma) \\
            &= \max(0,-v_{t,u} \, C_1)^2 + \max(0, -v_{t,u} \, C_2)^2.
\end{split}
\label{eq:HL2}
\end{equation}

The proposed loss function is the negative Sharpe ratio, combined with the variance-controlled loss terms

\begin{equation}
    L_1 = -SR_T + \xi \, \frac1T \left( \sum_{t=1}^T HL_1(\mathbf{w}_t; \gamma) \right),
    \label{eq:Proposed loss L1}
\end{equation}
and
\begin{equation}
    L_2 = -SR_T + \xi \, \frac1T \left( \sum_{t=1}^T HL_2(\mathbf{w}_t; \gamma) \right),
    \label{eq:Proposed loss L2}
\end{equation}
where $\xi$ is the parameter that controls the effects of the additional loss terms. We define the proposed methods as training a machine learning model that uses the proposed loss functions $L_1$ or $L_2$.


\section{Datasets and Evaluations}

In this section, we present empirical experiments on the baseline and proposed methods. Before that, details about the dataset, model architecture, feature selection, and hyperparameters setting are described.


\subsection{Dataset Description}
\label{Dataset Description}

We pulled the $1$-hour timeframe of OHLCV data of three assets with tickers name BTCUSD (BTC), BTCUPUSDT (BTCUP), and BTCDOWNUSDT (BTCDOWN) using Binance API. OHLCV is an aggregate form of market data standing for Open, High, Low, Close, and Volume. The entire dataset spans 19 months and starts from the introduction of BTCUP and BTCDOWN. The data is split following the forward validation split scheme, where the previous training and the test dataset are merged to form a new training dataset. The new test dataset is the data available right after the new training dataset. In our experiment, we use three test datasets. A summary of the data range in training and testing is shown in Table \ref{tab:data range}. The test periods are selected so that all have a duration of two months and try to cover different price trends of BTC. In the first and second periods, BTC has an uptrend price movement, whereas BTC lost about $25 \%$ of its value in the third period.

\begin{table}
    \centering
    \caption{Data range for training and testing.}
{\resizebox{\columnwidth}{!}{
    \begin{tabular}{lcc}
         \toprule
         
         & Train data range & Test data range \\
         
         \midrule 
         
         Period 1 & 2020-05-15 00:00 to 2021-07-03 23:00 & 2021-07-04 00:00 to 2021-09-01 23:00 \\
         Period 2 & 2020-05-15 00:00 to 2021-09-01 23:00 & 2021-09-02 00:00 to 2021-10-31 23:00 \\
         Period 3 & 2020-05-15 00:00 to 2021-10-31 23:00 & 2021-11-01 00:00 to 2021-30-12 23:00 \\
         
         \bottomrule
    \end{tabular}%
}}
    \label{tab:data range}
\end{table}



\subsection{Model Architecture and Feature Selection}
\label{Model architecture}

The input features for the neural network are derived from OHLCV data and return data for all the assets in the portfolio. Returns for individual assets are computed from the closing prices of two sequential periods, as defined in Eq. (\ref{eq:asset_return}).
Given the heterogeneity in price and volume ranges among assets, we apply normalization for all data, including returns. For each type of data, the mean and standard deviation in $L_{norm}$ consecutive periods are computed, and these statistics are then used to perform the z-score normalization for the successful $L_{norm}$ periods. This forwarding normalization scheme guarantees that information from the future does not flow backward in time.

\begin{figure}
    \centering
    \captionsetup[subfloat]{font=scriptsize}
    \subfloat[Feature selection\label{fig:Feature selection}]{\includegraphics[width=5cm]{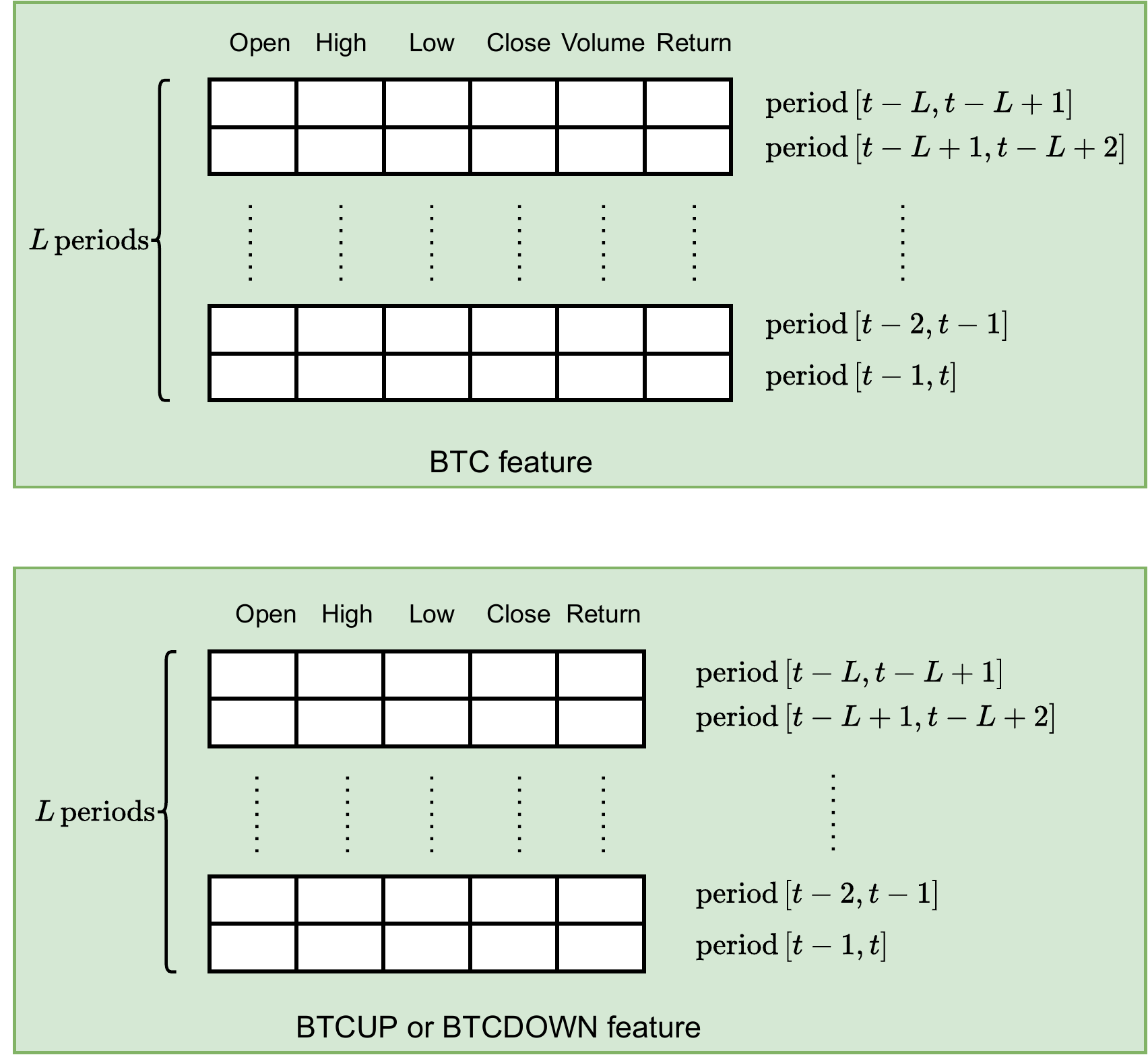}}
    \hfill
    \subfloat[Model architecture\label{fig:Model architecture}]{ \includegraphics[width=3.3cm]{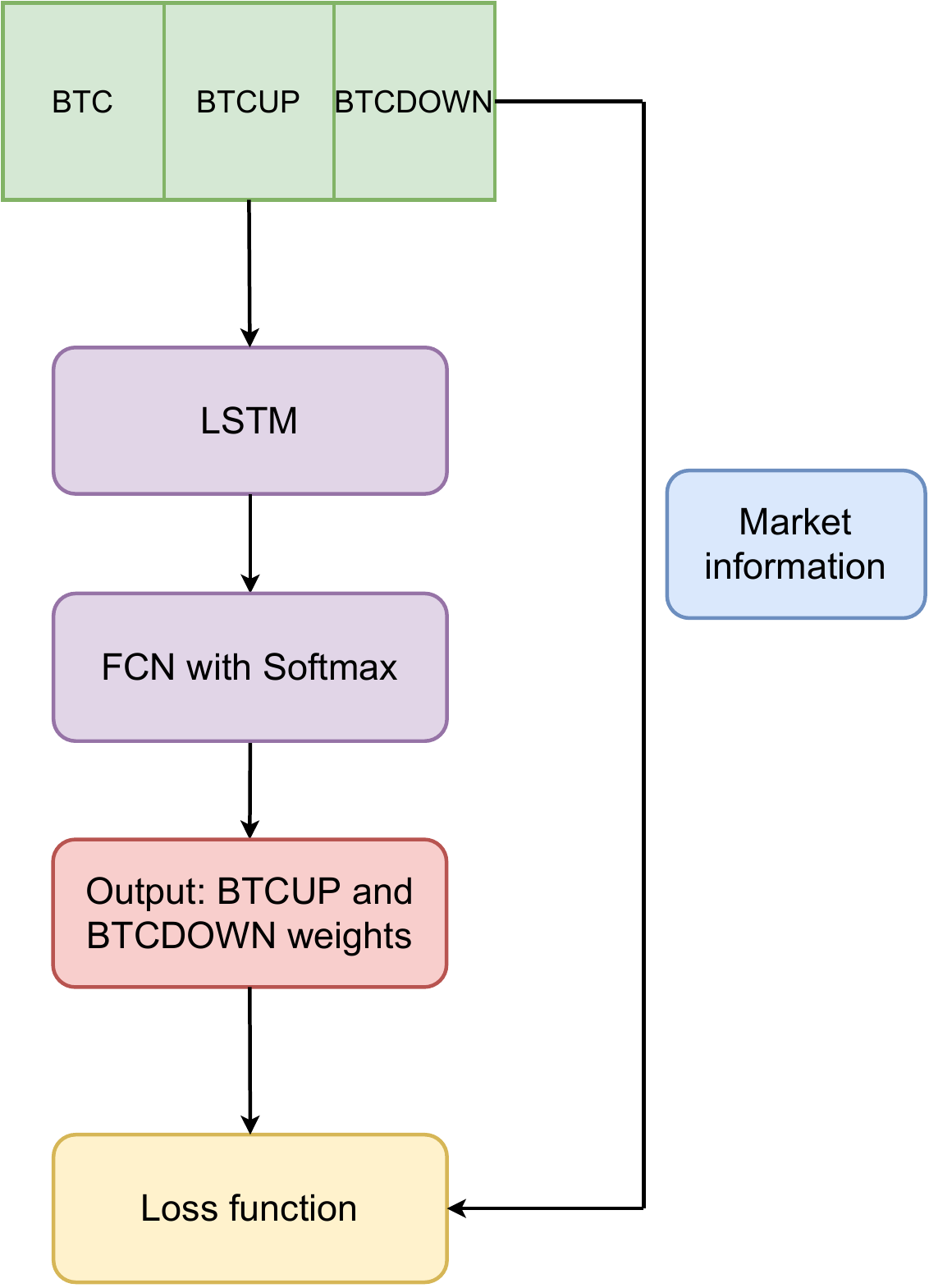}}
    \caption{In \protect\subref{fig:Feature selection}, the elements of feature matrices are the z-score normalization data. \protect\subref{fig:Model architecture} shows the forward computation of the model and its structure. The market information are the BTCUP and BTCDOWN close prices, their future returns, and $\beta_t^{market}$.}
    \label{fig:Feature selection and model architecture}
\end{figure}

After normalization, the feature of each ticker is organized into a matrix where the columns are concatenated from the OHLCV and return features. The features of the three assets are combined along the column dimension. Figure \ref{fig:Feature selection} visualizes the input feature. To estimate the portfolio weights $\mathbf{w}_t$, the model uses a lookback window containing data from the most recent $L$ periods, including period $[t-1,t]$.  

Our choice of network is a Long Short-Term Memory (LSTM) \cite{hochreiter_long_1997} with one layer and the hidden feature size of $64$, followed by a fully connected layer with the Softmax activation function. While there exist numerous specialized network architectures, such as \cite{tsantekidis2017forecasting, tran2017tensor, tran2018temporal, tran2019data, tran2022attention, tran2021data}, proposed for the purpose of financial forecasting, our choice leans towards LSTM (Long Short-Term Memory). This is primarily due to its robust validation through extensive testing in portfolio optimization \cite{zhang_deep_2020}. Figure \ref{fig:Model architecture} summarizes the model architecture.



\subsection{Experiment Protocols}
\label{Experiment Protocols}

We perform a hyperparameter search using the following parameters. The batch size is selected from the set $\{ 64, 128, 256, 512 \}$, while the number of epochs ranged between $80$ and $140$ in increments of $20$. We use Adam optimizer \cite{kingma_adam_2014} and a cosine annealing learning rate scheduler \cite{loshchilov_sgdr_2016} with the start learning rate chosen from the set $\{ 1\text{e-}5, 3\text{e-}5, 5\text{e-}5, 1\text{e-}4, 3\text{e-}4, 5\text{e-}4, 1\text{e-}3 \}$. The weight decay is chosen from $ \{ 0.0, 1\text{e-}4, 3\text{e-}4, 5\text{e-}4, 1\text{e-}3 \} $. The value of $\gamma$ is chosen from the set $ \{ 0.0, 0.1, 0.2, \dots, 1.0 \}$. The parameter $\xi$ is selected from the set $ \{ 1\text{e-}6, 3\text{e-}6, 5\text{e-}6, 1\text{e-}5, 3\text{e-}5, 5\text{e-}5, 1\text{e-}4, 3\text{e-}4, 5\text{e-}4, 1\text{e-}3 \} $. We conduct experiments under both fee-included and no-fee configurations. In the no-cost scheme, $\mu_t$ is set to $1$, and no management fee is imposed. Conversely, the cost-included scheme incorporated the Binance trading fee for BLVTs, setting the trading fee rate $c$ at $0.075 \%$, and the daily management fee $m$ at $0.01 \%$. Lastly, the length of the normalization window is fixed at $ L_{norm} = 12 $ and the number of periods used to estimate $\beta^{market}$ and the length of the lookback window is fixed at $L=K=48$.

Due to the random initialization of the model parameters, the performance of the trained models differs for each training session. Each parameter configuration is run five times, and the reported metric is the mean and standard deviation of the average Sharpe ratio over three testing periods across five runs.



\subsection{Experimental Results}
\label{Experimental Results}


We compare our proposed methods with the baseline method (NS). The proposed methods are presented under the name $\text{SVC}_1$ for model training with loss function $L_1$, and $\text{SVC}_2$ for loss function $L_2$. We also include other benchmark methods for a comprehensive comparison. The Neutral-Weight Portfolio (NWP) involves the allocation of estimated neutral weights as outlined in Eq. (\ref{eq:Delta-neutral weight estimation}). The Equal-Weight Portfolio (EWP) \cite{demiguel_optimal_2009} uniformly distributes weights between BTCUP and BTCDOWN for each allocation period. The Global Minimum Variance Portfolio (GMVP) \cite{markowitz_portfolio_1952} determines optimal weights that minimize the estimated return covariance matrix, using a window length of $48$ hours for estimation. Finally, the performance of the underlying cryptocurrency of BTCUP and BTCDOWN, which is BTC, will be presented for reference. For holding BTC, the performance when the trading fee and management fee are included does not affect the portfolio performance since we do not conduct any trading, and holding BTC is not required the management fee.

\begin{table*}
\caption{Average Sharpe ratio over three testing periods.}

    \centering
    \begin{tabular}{lccccccccc}
    
    \toprule
    
    & NS & $\text{SVC}_1$ (proposed) & $\text{SVC}_2$ (proposed) & NWP & EWP & GMVP & BTC \\ 

    \midrule
        No fee & 0.033 $\pm$ 0.008 & \bfseries 0.035 $\pm$ 0.005 & 0.033 $\pm$ 0.004
        & -0.005 & 0.004 & -0.001 & 0.012 \\
         Fee included & 0.030 $\pm$ 0.007 & \bfseries 0.033 $\pm$ 0.003  & 0.032 $\pm$ 0.002
         & -0.008 & -0.003 & -0.010 & 0.012 \\
    \bottomrule
    
    \end{tabular}%
    \label{tab:average sharpe ratio}
\end{table*}

\begin{table}
    \caption{Experiment results with no trading fee and management fee.}
\resizebox{\columnwidth}{!}{%
    \centering
    \begin{tabular}{lccc|ccc|ccc}

        \toprule

       & \multicolumn{3}{c|}{Period 1} & \multicolumn{3}{c|}{Period 2} & \multicolumn{3}{c}{Period 3} \\
       & Sharpe & fAPV & MDD & Sharpe & fAPV & MDD & Sharpe & fAPV & MDD \\
       
       \midrule
       
       NS  & 0.040 & 1.338 & 0.147 &
           0.034 &  1.385 & 0.221 &
          \bfseries 0.029 & \bfseries 1.258 & 0.218 \\

        $\text{SVC}_1$ & \bfseries 0.079 & \bfseries 1.878 & 0.142 &
                 0.033 & 1.401 & 0.220 &
                 -0.010 & 0.855 & 0.398 \\
         
       $\text{SVC}_2$ & 0.056 & 1.299 & 0.060 &
         \bfseries 0.043 & \bfseries 1.438 & 0.167  &
         0.002 & 0.973 & 0.329 \\
                
        NWP & 0.003 & 1.007 &  0.050 &
            -0.016 & 0.956 &  \bfseries 0.053 &
            -0.001 & 0.995 & 0.053 \\
                
        EWP & -0.006 & 0.983 & 0.077 & 
            -0.009 & 0.972 & 0.088 &
            0.026 & 1.057 & 0.038 \\
        
        GMVP & -0.004 & 0.991 & \bfseries 0.038 &
            -0.017 & 0.958 & 0.054 &
            0.019 & 1.035 &  \bfseries 0.033 \\
            
        BTC & 0.041 & 1.419 & 0.156 &
            0.024 & 1.220 & 0.231 &
            -0.028 & 0.754 & 0.335 \\
    
    \bottomrule

    \end{tabular}%
}
    \label{tab:result breakdown, no cost and fee}
\end{table}

The findings presented in Table \ref{tab:average sharpe ratio} indicate that our proposed methods enhance the performance of the baseline method in most cases. Other strategies such as NWP, EWP, and GMVP yield Sharpe ratios near zero, underscoring the advantages of deploying neural network-based methods over conventional ones. The performance of BTC is much lower than the performance of neural network-based methods because the profits of BTC gains in the first and second periods are eroded by the downtrend price movement in the third period, while the neural network-based methods can put more weight on BTCDOWN to speculate on the decreasing BTC price. Therefore, these results show the benefit of holding a pair of BLVTs over the underlying asset.

In Tables \ref{tab:result breakdown, no cost and fee}, and \ref{tab:result breakdown, cost and fee}, we choose the median result in the $5$ runs to show a breakdown of the performance of all methods over three testing periods to observe how each method performs in different market situations. In addition, the final accumulated portfolio value (fAPV) is presented as a profitability metric. We set the initial portfolio value to 1. Then the fAPV will reflect the accumulated return over each testing period. The Maximum Drawdown (MDD) is also considered to highlight the risk associated with each method.

\begin{table}[t]
    \caption{Experiment results with trading fee $c=0.075\%$ and daily management fee $m=0.01\%$}
\resizebox{\columnwidth}{!}{%
    \centering
    \begin{tabular}{lccc|ccc|ccc}

        \toprule

       & \multicolumn{3}{c|}{Period 1} & \multicolumn{3}{c|}{Period 2} & \multicolumn{3}{c}{Period 3} \\
       & Sharpe & fAPV & MDD & Sharpe & fAPV & MDD & Sharpe & fAPV & MDD \\ 
    
    \midrule
    NS & 0.075 & \bfseries 1.939 & 0.097 &
        0.017 & 1.189 & 0.311 &
        -0.005 & 0.848 & 0.472 \\
        
  $ \text{SVC}_1$ & 0.074 & 1.891 & 0.123 &
            0.025  & 1.319 & 0.319 &
        \bfseries 0.003 & \bfseries 0.995 & 0.214 \\
         
    $\text{SVC}_2$ & \bfseries 0.075 & 1.862 & 0.113 &
           \bfseries 0.026 & \bfseries 1.324 & 0.323 &
            -0.005  & 0.868  & 0.459\\
            
    NWP & 0.000 & 0.998 & 0.050 &
        -0.019 & 0.948 & \bfseries 0.056 &
        -0.005 & 0.987 & 0.056 \\
            
    EWP & -0.012 & 0.965 & 0.087 &
            -0.015 & 0.954 & 0.094 &
            0.018 & 1.038 & 0.039 \\
    
    GMVP & -0.013 & 0.973 & \bfseries 0.046 &
        -0.025 & 0.940 & 0.069 &
        0.009 & 1.016 & \bfseries 0.034 \\
        
  BTC & 0.041 & 1.419 & 0.156 &
        0.024 & 1.220 & 0.231 &
        -0.028 & 0.754 & 0.335 \\
    \bottomrule

    \end{tabular}%
}
    \label{tab:result breakdown, cost and fee}
\end{table}


These tables show that the NWP, EWP, and GMVP methods usually have the lowest MDD in all test periods, whether the trading and management fee are considered. While these methods are effective in reducing portfolio risk, they do so at the expense of portfolio profitability. The fAPVs of these methods show that they can only preserve the original portfolio value and cannot make profits. The Sharpe ratios from the proposed methods outperform the baseline method in mostly every period in different settings. When BTC loses one-fourth of its value in the third period, the proposed methods can profit and achieve a higher Sharpe ratio than both BTC and the baseline method.






\vfill\eject

\section{Conclusion}

This work adopts the approach of using deep learning for the portfolio optimization problem with the Sharpe ratio as the loss function. A pair of BLVTs is chosen to construct a portfolio to benefit from their consistent negative correlation. The portfolio contains only two assets that allow us to investigate the neutral position of the portfolio using simple mathematical analysis. Additional loss terms are designed to control the neutrality of the portfolio. We compare the proposed methods with the baseline and other non-learning approaches. Experimental results show that holding a portfolio containing a pair of BLVTs is superior to holding only the underlying asset, especially for a high-volatility market like cryptocurrency. In addition, the proposed methods show their effectiveness in improving the baseline method and gaining the best results compared to other methods in all different settings.



\bibliographystyle{IEEEtran}
\bibliography{IEEEabrv,references.bib}

\end{document}

%% file: glossary.tex
\newacronym{ai}{AI}{Artificial Intelligence}

\newacronym{btc}{BTC}{Bitcoin}

\newacronym{rnn}{RNN}{Recurrent Neural Network}

\newacronym{cnn}{CNN}{Convolutional Neural Network}

\newacronym{blvts}{BLVTs}{Binance Leverage Tokens}

\newacronym{fAPV}{fAPV}{final accumulated portfolio value}

\newacronym{mdd}{MDD}{maximum drawdown}

\newacronym{nav}{NAV}{Net Asset Value}

\newacronym{lstm}{LSTM}{Long Short-Term Memory}

\newacronym{fcn}{FCN}{Fully connected neural network layer}

\newacronym{dn}{DN}{Delta-neutral portfolio}

\newacronym{ew}{EWP}{equal-weighted portfolio}

\newacronym{tabl}{TABL}{Temporal Attention augmented Bilinear Layer}

\newacronym{rl}{RL}{Reinforcement Learning}

\newacronym{gmvp}{GMVP}{Global Minimum Variance portfolio}

\newacronym{dpm}{DPM}{Deep Portfolio Management}

\newacronym{eiie}{EIIE}{Ensemble of Identical Independent Evaluators}

\newacronym{hrpm}{HRPM}{Hierarchical Reinforced trading system for Portfolio Management}